\documentclass[conference]{IEEEtran}
\IEEEoverridecommandlockouts
\usepackage{cite}
\usepackage{amsmath,amssymb,amsfonts}
\usepackage{algorithmic}
\usepackage{graphicx}
\usepackage{textcomp}
\usepackage{xcolor}
\usepackage{graphicx}
\usepackage{subcaption}
\usepackage{listings}
\usepackage{balance}

\usepackage{url}
\usepackage{xspace}%

\newcommand{\eg}{e.g.,\xspace}

\newcommand{\ie}{i.e.,\xspace}

\newcommand{\quotes}[1]{``{#1}''}
\newcommand{\quoquo}[1]{`{#1}'}
\newcommand{\sd}{\textit{SD}}
\newcommand{\me}{\textit{M}}

\def\BibTeX{{\rm B\kern-.05em{\sc i\kern-.025em b}\kern-.08em
    T\kern-.1667em\lower.7ex\hbox{E}\kern-.125emX}}
\begin{document}

\title{Applying General Turn-taking Models to Conversational Human-Robot Interaction}

\author{\IEEEauthorblockN{Gabriel Skantze}
\IEEEauthorblockA{\textit{Division of Speech, Music and Hearing} \\
\textit{KTH Royal Institute of Technology}\\
Stockholm, Sweden \\
skantze@kth.se}
\and
\IEEEauthorblockN{Bahar Irfan}
\IEEEauthorblockA{\textit{Division of Speech, Music and Hearing} \\
\textit{KTH Royal Institute of Technology}\\
Stockholm, Sweden \\
birfan@kth.se}
}

\maketitle

\begin{abstract}
Turn-taking is a fundamental aspect of conversation, but current Human-Robot Interaction (HRI) systems often rely on simplistic, silence-based models, leading to unnatural pauses and interruptions. This paper investigates, for the first time, the application of general turn-taking models, specifically TurnGPT and Voice Activity Projection (VAP), to improve conversational dynamics in HRI. These models are trained on human-human dialogue data using self-supervised learning objectives, without requiring domain-specific fine-tuning. We propose methods for using these models in tandem to predict when a robot should begin preparing responses, take turns, and handle potential interruptions. We evaluated the proposed system in a within-subject study against a traditional baseline system, using the Furhat robot with 39 adults in a conversational setting, in combination with a large language model for autonomous response generation. The results show that participants significantly prefer the proposed system, and it significantly reduces response delays and interruptions.
\end{abstract}


\section{Introduction}

Turn-taking is one of the most fundamental aspects of conversation. Since it is difficult to speak and listen at the same time, the speakers need to coordinate who is currently speaking, and when the turn should shift to the other person \cite{duncan1972some,sacks1978simplest}. Traditionally, conversational systems, including those for Human-Robot Interaction (HRI), have relied on a simplistic model of turn-taking based on heuristics, where the system waits for a certain amount of silence in the user's speech before deciding to take the turn, and only then starts to process the user's speech and finally generate a response \cite{skantze2021review}. However, since silence alone is not a reliable indicator that a user is yielding the turn (they may simply pause mid-sentence), this approach often results in either long response delays or frequent interruptions, depending on the set silence threshold \cite{majlesi_managing_2023}. Additionally, because it is difficult to differentiate genuine interruptions from collaborative overlapping speech, such as \textit{backchannels}  (\eg \quotes{mhm}, \quotes{yeah}), many systems avoid processing user input while the system is speaking. 

This is very different from the sophisticated coordination of turn-taking in human-human dialogue, where gaps between turns may be as brief as 0.2 s \cite{Levinson2015TurnTaking}. Humans rely on a number of different coordination cues to determine whether the other speaker is \textit{holding} or \textit{yielding} the turn, including intonation \cite{local100968,GravanoTT}, fillers (\eg \quotes{uh}, \quotes{uhm}), syntactic completeness \cite{ford100971}, gestures \cite{kendrick2023turn}, and gaze \cite{argyle101192,kendon100127}. Furthermore, the listener does not just wait for a signal to take the turn, they continuously try to predict (or `project') when the turn is about to end, to prepare their response in advance \cite{sacks1978simplest,Levinson2015TurnTaking,garrod2015use}. 

Although there are no systems today that can replicate this impressive coordination, several studies have shown that it is possible to develop more sophisticated, data-driven models of turn-taking in spoken dialogue systems \cite{raux2012,Meena2014,Lin2022} and HRI \cite{Johansson2015,Johansson2016,lala2019icmi}. However, these models are often trained on a specific turn-taking problem, using data specific to a particular domain, typically collected in similar settings. In many cases, collecting and annotating the data required for training in each new domain is impractical, especially for HRI.

In this paper, we explore the use of \textit{general} turn-taking models for HRI. By `general', we refer to models trained on larger sets of human-human conversational data using self-supervised learning objectives (thus avoiding annotation), without domain-specific fine-tuning. It also means that the models are generally applicable to identify a broad set of turn-taking events, including turn-yielding, turn-holding, backchannels, and interruptions. We specifically use two different models in tandem: TurnGPT \cite{ekstedt2020turngpt} and Voice Activity Projection (VAP) \cite{erik2022vap}. TurnGPT is a text-based model that incorporates the syntactic and semantic aspects of turn-taking, including more long-term pragmatic dependencies in a conversation. However, it does not consider prosodic or timing aspects. VAP, on the other hand, is trained purely on audio data, to continuously predict the dynamics of the conversation. 

Although previous studies have demonstrated that these models are effective at predicting turn-taking in human-human interactions, their training objective is to forecast what is likely to occur in the conversation, rather than optimizing specific behaviors for an agent. Using such a model to guide the behavior of a conversational agent (such as a robot) is not trivial. In this paper, we propose a method for achieving this within an HRI context. We evaluated the proposed system in a within-subject study, comparing it to a traditional baseline, utilizing the Furhat robot \cite{AlMoubayed2012} with 39 adult participants.

Evaluating such models in an HRI setting with users is important, since more sophisticated turn-taking model might not necessarily provide a better experience. These general models are trained to model the speech activity of humans with certain conversational dynamics and characteristics. In an HRI setting, these dynamics might be different, and the speech from the robot is not natural but synthesized. Furthermore, the HRI setting involves additional cues, such as gaze, and may be affected by users adapting their communication style depending on their mental model of their conversation partner (\ie a robot)~\cite{gallois2005communication}. If the input to the model is out-of-distribution, it might result in bad predictions and, thus, bad input to the robot's control system. It is also not certain that humans would prefer to interact with a robot that is more human-like, and might instead prefer a more traditional (and perhaps more predictable) turn-taking mechanism. 

\section{Related Work}

There are several aspects of turn-taking that a conversational agent should consider. One of the key challenges is distinguishing when the user is holding or yielding the turn, allowing the agent to respond quickly when the turn is yielded and avoid interrupting during pauses. The moments when turn shifts are likely to occur are often called `Transition-Relevant Places' (TRPs) \cite{sacks1978simplest}. 
Several studies have explored how machine learning can address this challenge by analyzing verbal and non-verbal signals \cite{ferrer100384,GravanoTT,maier_towards_2017,skantze2017sigdial,Onishi2023,shahverdi_learning_2022}. However, much of this work reports only prediction performance on corpora, which does not necessarily translate to real-world performance, where systems must interact with users in real-time and face processing constraints. That said, such models have been applied and evaluated within spoken dialogue systems \cite{raux2012,Meena2014,Lin2022} and in human-robot interaction (HRI) contexts \cite{Johansson2016,lala2019icmi,yang_gated_2022}. While some HRI settings are dyadic, they can also involve multiple users, which makes the problem more challenging, as the robot also needs to determine whether the users are addressing the robot or each other \cite{Johansson2015,Tesema2023}. 

Another turn-taking challenge is identifying moments when it is appropriate to produce backchannels \cite{skantze2021review}. Studies have demonstrated how this can be done both offline with corpora \cite{morency101537,Ruede2019} and online in HRI settings \cite{Park2017,murray_learning_2022}. Backchannel-inviting cues are similar, though not identical, to TRP cues  \cite{GravanoTT}, and backchannels more commonly occur in overlaps \cite{yngve100128}.

While dialogue systems can be implemented using a \textit{simplex} channel (where only one interlocutor can speak at a time), it is often desirable to have a \textit{duplex} channel, enabling the system to listen while speaking (to hear overlapping speech) \cite{Lin2022}. The most common use case for this is to allow the user to `barge-in' and interrupt the system mid-speech. However, a common issue with barge-in is false triggers, caused either by external noise or coughing, or by the user offering a backchannel without intending to take the turn. If the system stops speaking abruptly in such cases, it can lead to confusion, especially if it cannot resume seamlessly. For duplex systems, it is therefore essential to have a model that can distinguish genuine interruptions from collaborative overlapping speech or other sounds \cite{Lee2010,oertel101561,Truong2013,cumbal_let_2024}. If not handled appropriately, allowing for user interruptions might be detrimental to the user experience \cite{heins100496,Rose2003,skantze2021review}, and a simplex system can be preferable. 

Our work is the first to demonstrate how general turn-taking models can be used to address all these challenges, when applied to an HRI setting, rather than relying on separate models for each. Two other aspects also make our approach novel. First, while many models only make predictions at specific events (such as when the user becomes silent) \cite{raux2012,Meena2014}, our models are \textit{continuous}, providing output at every time step. This enables the system not only to take turns appropriately but also to plan responses in advance, resolve interruptions, and handle backchannels effectively. Second, our models utilize \textit{self-monitoring}, meaning they account for the robot's speech as well as the user’s. This is critical because the robot’s speech provides context for understanding the user’s turn-taking cues and allows the system to \quotes{reflect} on potential cues in its own speech \cite{ekstedt23tts}. 

\section{General Turn-taking Models}
\label{sec:predictive-models}

In this paper, we use two general turn-taking models: TurnGPT and VAP that have complementary properties. Both models are trained on general human-human dialogue datasets in a self-supervised fashion. This means that they do not require any manual annotation of the data and can, therefore, be trained on large datasets. The models have previously been evaluated offline with human-human conversation datasets that are either text or speech-based, but have not been applied to an HRI setting. 

\begin{figure*}[t!]
    \centering
    \includegraphics[width=\textwidth]{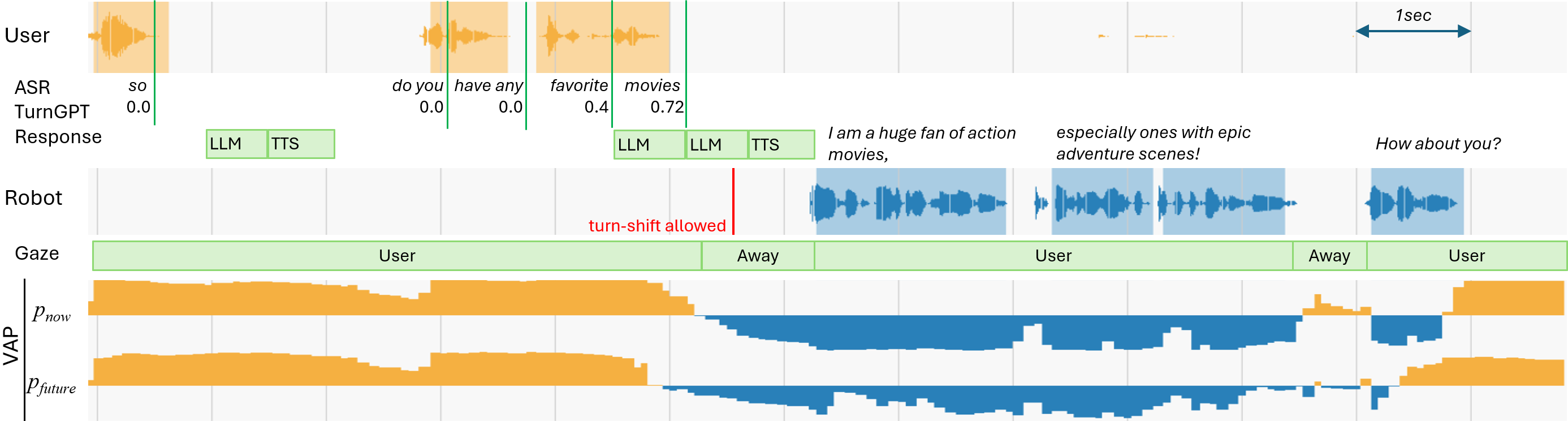}
    \caption{Example turn shift from user to robot when using the proposed system. From top to bottom: (1) The user's actual speech pattern (in orange), highlighted where the VAP model detects voice activity; (2) ASR transcription; (3) TurnGPT's likelihood for the turn to end; (4) Response generation (LLM+TTS); (5) the robot's actual speech pattern (in blue); (6)  robot gaze (towards user or averted); (7) VAP model predictions.}
    \label{fig:vap_turnshift}
\end{figure*}

\subsection{TurnGPT: Predictions in the Verbal Domain}

\newcommand{\ts}{\scalebox{0.7}[1]{\texttt{\textless ts\textgreater}}}

In the verbal domain, a clear cue for turn-holding is when an utterance is syntactically incomplete. Obvious instances, such as \quotes{I would like to order a ...} can be readily identified through part-of-speech tagging \cite{GravanoTT,Meena2014}. However, in ongoing dialogue, it is often more relevant to consider \textit{pragmatic} completeness, which takes into account the dialogue context \cite{ford100971}. For example, in the exchange \quotes{A: When will you leave? B: Tomorrow}, B's response is pragmatically complete, but only when considered in relation to A's preceding question.

TurnGPT \cite{ekstedt2020turngpt} is a model of syntactic/pragmatic turn completion, based on the textual representation of the dialogue. The model is an extension of GPT-2 \cite{gpt2}, where a special turn completion token, \ts{}, is included at the end of each turn in the training data. Similar to any GPT-based language model, it predicts the next token through a probability distribution over all tokens in the vocabulary, including \ts{}. The probability assigned to \ts{} can thus be regarded as a turn completion probability. The model used in this paper was trained on the SODA dataset of 385K text-based conversations \cite{kim2022soda}. TurnGPT responds in about 20ms (with the GPU used in this work: 8GB GeForce RTX 2080), which is not feasible to achieve with bigger models to respond in real time.

Previous work has shown that TurnGPT can model long-term pragmatic dependencies over several turns \cite{ekstedt2020turngpt}. 
However, TurnGPT does not model any temporal or prosodic aspects of the speech signal, and syntactic/pragmatic completeness alone is not always sufficient to identify the end of a turn. For example, the utterance \quotes{I would like to order a burger ... with some some fries ... and a milkshake.}, has several potential completion points. To know whether they constitute actual turn-yielding points, non-verbal cues are needed.

\subsection{VAP: Predictions in the Acoustic Domain}
\label{sec:predictive-models:vap}

Voice Activity Projection (VAP) \cite{erik2022vap} is a transformer-based model \cite{transformer} of conversational dynamics based purely on acoustic input. The model objective is to continuously (e.g., 10 times per second) predict (or `project') the upcoming voice activity of both speakers in a dialogue, in a 2-second future time window, based on the past 30 s of spoken dialogue. The model learns to predict complex turn-taking phenomena, such as turn-shifts, backchannels, and resolution of interruptions \cite{erik2022vap}. The model input is a raw waveform, which does not have the delay associated with models requiring speech recognition output (such as TurnGPT). In this paper, we use a version of the model \cite{ekstedt2023thesis} that takes in a stereo waveform (one channel for the user's speech and one for the robot's own speech). 

In this work, we used a VAP model trained on subsets of the Fisher Part 1 and Switchboard corpora \cite{fisher,swb}, both of which consist of recorded telephone conversations between U.S. speakers, totaling approximately 1,000 hours of dialogue. These corpora are also gender-balanced.

Following \cite{ekstedt2023thesis}, we use a simplified representation of the output, resulting in two values: $p_{now}$, which predicts the most likely speaker in the next 0-600 ms window, and $p_{future}$, which predicts the most likely speaker in the 600-2000 ms window following that. 

An example of these predictions, when the model is applied to HRI, can be seen in Figure~\ref{fig:vap_turnshift}. 
In the first pause, after the initial \quotes{so ...}, both $p_{now}$ and $p_{future}$ predict that the user will continue. When the user's turn is about to end (\quotes{...favorite movies}), both predictions shift to the robot, already before the robot has started to speak, where $p_{future}$ reacts slightly earlier than $p_{now}$. When the robot makes a pause towards the end (before \quotes{How about you?}), $p_{now}$ predicts a potential short speech activity from the user, while $p_{future}$ seems to favor the robot continuing to speak. This can be interpreted as an invitation to give a brief response, such as a backchannel. 

Since the VAP model is trained end-to-end on raw audio, it is not clear what turn-taking cues it has learned to pick up. However, previous work has shown that the model is sensitive to subtle prosodic cues \cite{erik2022sigdial} and to fillers \cite{jiang2023}. While it is also possible that the model has learned to pick up certain verbal cues, even though there is no explicit speech-to-text objective involved, it is not likely that it can make use of more long-term pragmatic completion cues (which TurnGPT can). 

While TurnGPT and VAP offer complementary capabilities, they cannot be easily combined. The number of conversations in the spoken dataset (about 10K) used to train VAP is not enough to train an LLM such as TurnGPT (which was trained with 385K text-based conversations), and the text datasets cannot be used to train a speech model like VAP. 
We therefore propose to use VAP and TurnGPT together, in tandem. 

\section{Baseline System}

To apply the proposed system to HRI, we used the Furhat robot \cite{AlMoubayed2012}, a human-like robot head featuring an animated face projected onto a semi-translucent mask, along with a mechanical neck, as seen in Figure~\ref{fig:setting}. This robot platform was selected for its expressiveness, which includes facial expressions, lip movements, gaze behavior, and head gestures. 

As a baseline to compare our proposed turn-taking system with, we use the standard turn-taking mechanism used in the Furhat software. This mechanism is very similar to most traditional spoken dialogue systems, including many deployed systems \cite{skantze2021review}. 
In order to make this version as good as possible, given the absence of any turn-taking model, we incorporated additional elements, such as gaze aversion and a LED signal, to offer turn-taking cues that allow for a more balanced comparison with our proposed model.

\begin{figure}[t]
    \centering
    \includegraphics[width=\columnwidth]{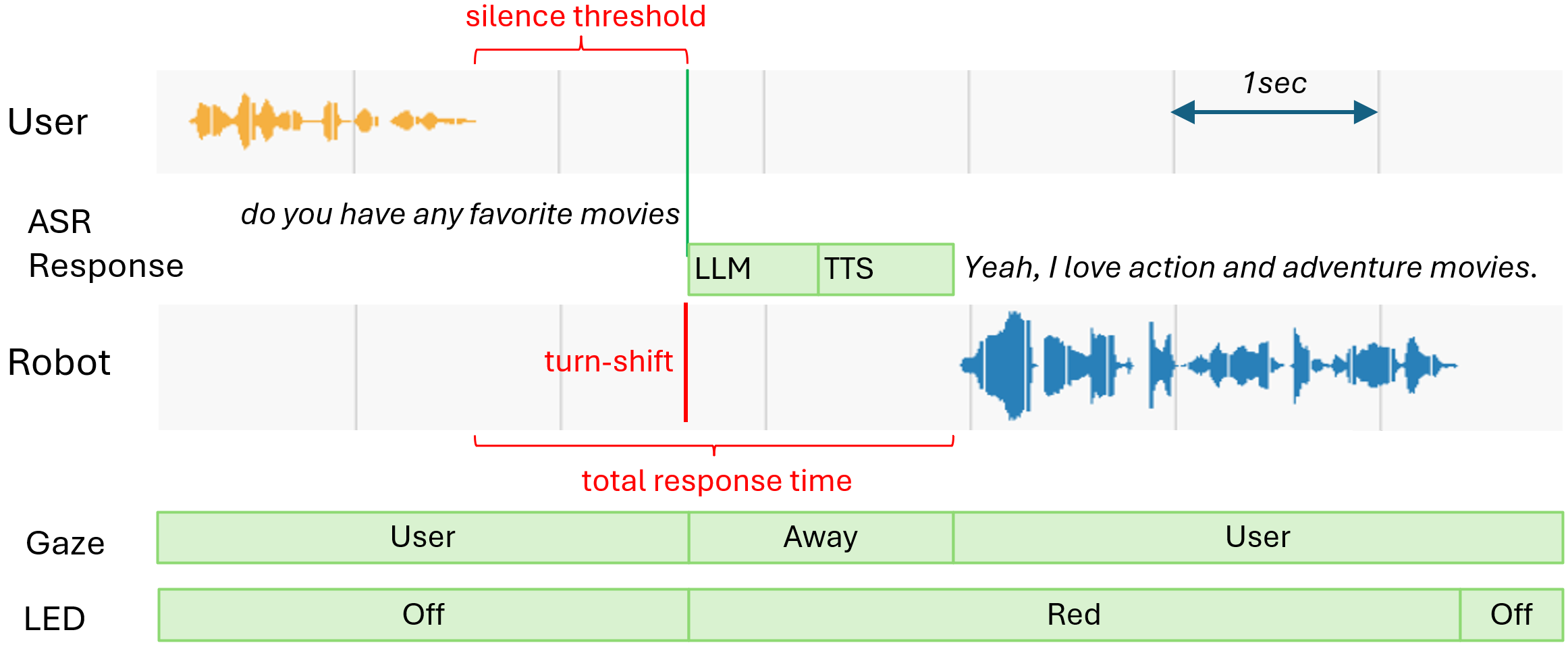}
    \caption{Example turn shift from user to robot with the baseline.}
    \label{fig:baseline}
\end{figure}

\subsection{Baseline Turn-taking Mechanism}

Figure~\ref{fig:baseline} illustrates a turn-shift from user to robot. For Automatic Speech Recognition (ASR), we use Google's Speech-to-Text service. Audio from the microphone is streamed to the ASR, which also determines the end-of-speech. Although the algorithm behind Google's end-of-speech detection is not documented, it is likely based on a silence threshold, since it is fairly constant in length. At the end-of-speech, the speech recognition result is used to build a large language model (LLM) prompt, which is sent to OpenAI gpt-3.5-turbo, hosted by Microsoft Azure.

The result from the LLM (the robot's response) is sent to the text-to-speech (TTS), and the audio is played back to the user. In order to have a human-like robot voice to allow for a natural HRI, ElevenLabs\footnote{\url{https://elevenlabs.io/text-to-speech}} TTS was used, with the voice \quoquo{Jennifer}. We used the streaming version of the ElevenLabs API, resulting in a fairly constant processing delay (around 1 s), regardless of the length of the utterance to produce. As visible in Figure~\ref{fig:baseline}, the total response time (measured from the actual point where the user stopped speaking until the robot starts speaking) is almost 2.5 s in this example, though it can vary depending on the processing delay of the LLM and TTS.

The standard Furhat software does not support user interruptions (`barge-in'). 
As discussed earlier, allowing for user interruptions can easily lead to confusion if not modeled appropriately. Thus, in the absence of any turn-taking model for the baseline system, we followed this design choice. However, to make it clear to the user when the robot was listening, similar to \cite{irfan2023between}, we implemented both a gaze aversion mechanism (see next section), and used the robot's LED at the base to signal its listening status, turning it red when not listening, and off otherwise, as illustrated in Figure~\ref{fig:baseline} and seen in Figure~\ref{fig:setting}. This was clearly communicated to the participants during evaluation through a demonstration video.

\subsection{Non-verbal Robot Behavior}
\label{sec:gaze-aversion}

We adopted a gaze aversion strategy as a turn-taking signal in which the robot looks at a random diagonal away from the user when it starts to generate a response after identifying a turn shift, as shown in Figure~\ref{fig:baseline}. In addition, we added gaze aversion at every pause within the robot's utterance to signal that the turn had not yet been yielded.  

To improve the consistency of the responses, 
we used dynamic facial expressions with an LLM, similar to \cite{mishra_real-time_2023}. At the onset of the robot's speech, numbered anchor points were inserted in the text at phrase boundaries. LLM was then asked to insert suitable facial expressions at each anchor point, based on the list of available expressions in the Furhat SDK. This was done asynchronously so as not to add any further delays. 

\section{Proposed Turn-taking System}

\begin{figure}[h!]
    \centering
    \includegraphics[width=\columnwidth]{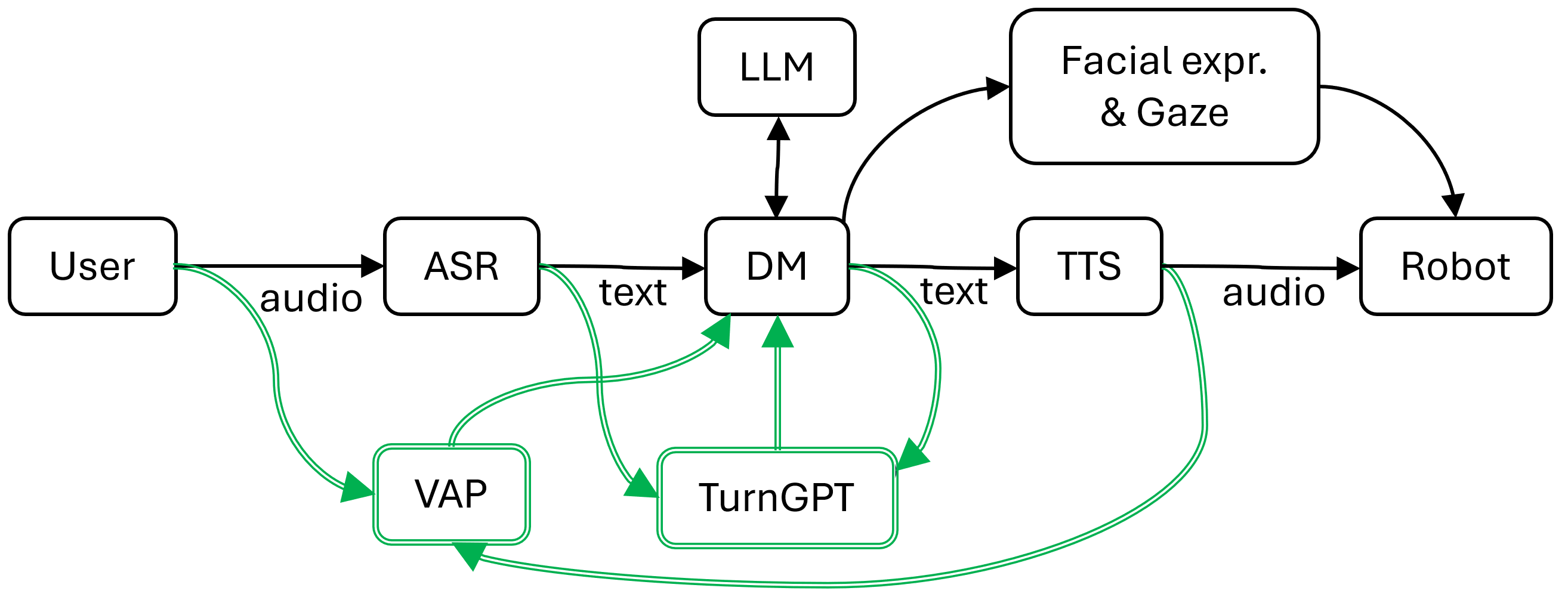}
    \caption{System architecture. New components in proposed system shown in green.}
    \label{fig:system}
\end{figure}

In the proposed system, we use the two general turn-taking models described in Section~\ref{sec:predictive-models}. 
The models are trained to make predictions as third-party observers of a conversation between two speakers. Thus, it is not enough to include these models on the input side of the dialogue system, we also need to feed the TTS output (audio) back to the VAP model and the resulting dialogue history to TurnGPT, as shown in Figure~\ref{fig:system}. This `self-monitoring' is an important new aspect of the dialogue system architecture, as traditional spoken dialogue systems typically never analyze their own spoken output. 

Since the models are not trained to optimize a certain behavior, and we are not fine-tuning the models against some domain-specific dataset, but rather use them in a `zero-shot' fashion, the behavior of the agent will depend on a number of (fairly arbitrary) hyper-parameters. Their current values were tuned through four pilot experiments, but they can likely be optimized further in future work. The complete pseudo-code for the turn-taking algorithm and the values for the hyper-parameters are given in Appendix A.

\subsection{Identifying the End of the User's Turn}

Figure~\ref{fig:vap_turnshift} shows how a user-robot turn shift is handled by the proposed system. We use the streaming results from the ASR, which are sent to TurnGPT, and the audio from the two channels are analyzed by VAP in real time. The output of the two models is then used to determine whether the user has yielded the turn or not. After the initial \quotes{So...}, TurnGPT assigns a probability of 0.0 for a turn shift, and the VAP model favors the user throughout the long pause. With the baseline system, the user would have been interrupted since the silence length threshold would have been triggered. TurnGPT then continues to assign low turn-shift probabilities until the words \quotes{favorite} and \quotes{movies}, which both constitute potential completion points. Here, the VAP model also indicates the end of the turn. After both $p_{now}$ and $p_{future}$ have favored the user for at least 0.5 s, a turn-shift is \textit{allowed} (which does not necessarily mean that it is ready to speak yet). 

Neither of the models makes perfect predictions. The VAP model, for example, sometimes fails to identify turn yielding moments. Thus, even if VAP continues to favor the user, a timeout is used to determine when the robot is allowed to start speaking, depending on the turn-shift probability from TurnGPT, where the maximum timeout was set to 3 s.

In ideal circumstances, this algorithm allows the user to pause for up to 3 s, while the robot may be able to take the turn with just 0.5 s response delay. 

\begin{figure*}[t]
    \centering
    \begin{minipage}[t]{0.6\textwidth}
        \centering
        \includegraphics[width=\textwidth]{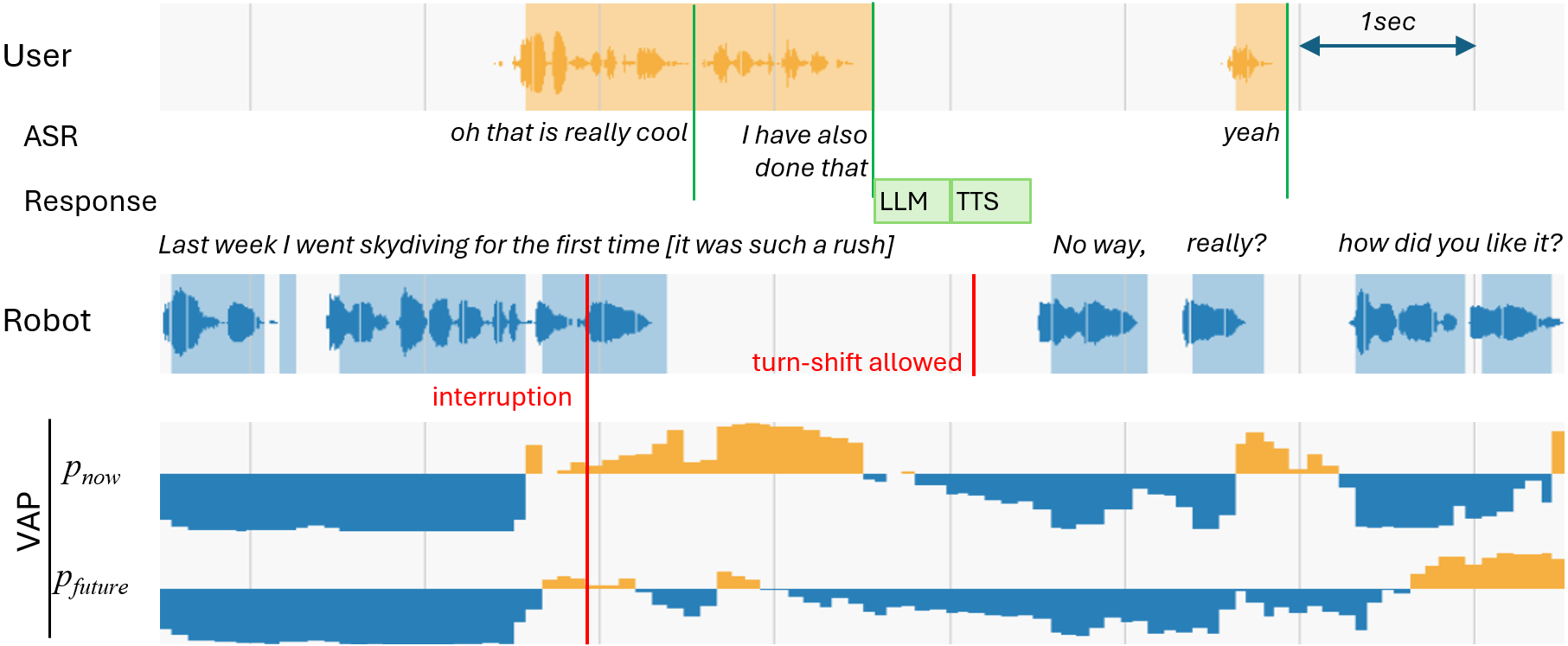}
        \caption{Example of a user interruption and backchannel.}
        \label{fig:interruption}
    \end{minipage}%
    \hfill
    \begin{minipage}[t]{0.37\textwidth}
        \centering
        \includegraphics[width=\textwidth]{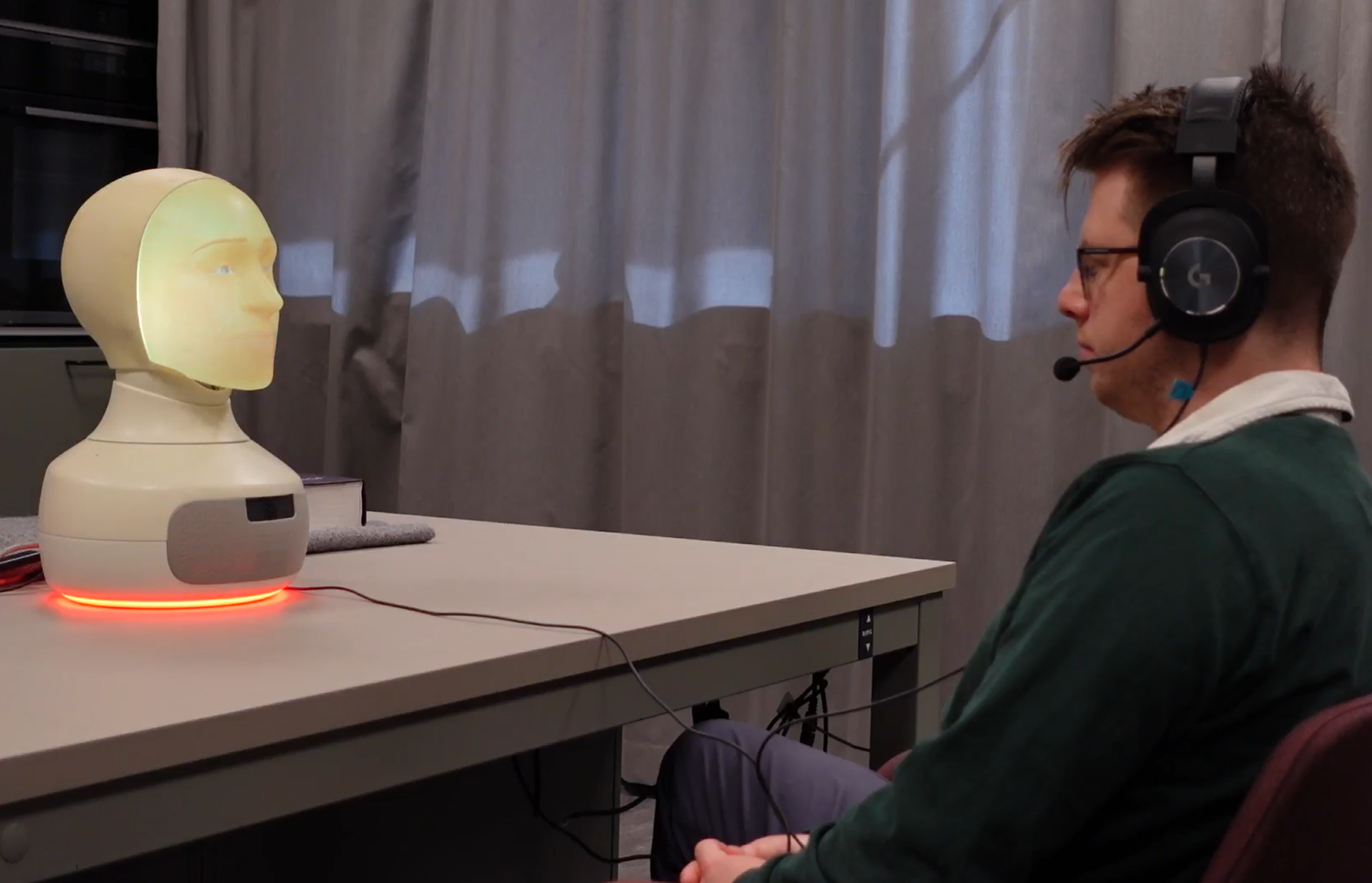}
        \caption{The setting for the evaluation, showing the red LED lights used in the baseline condition to indicate that the robot is not listening.}
        \label{fig:setting}
    \end{minipage}
\end{figure*}

\subsection{Preparing a Response}

While the system can respond 0.5 s after the user has finished speaking, this is typically not enough time to receive the most complete ASR result, 
as well as the results from the LLM and TTS. Similar to how humans manage turn-taking \cite{sacks1978simplest,Levinson2015TurnTaking,garrod2015use}, the system should start to prepare a \textit{tentative} response before the interlocutor is done speaking. 

While it is possible to start preparing a tentative response for each new incremental ASR result, we try to reduce compute by only doing this if either the turn-shift probability (according to TurnGPT) is 0.2 or higher, \textit{or} 0.2 s has passed since the previous incremental result. This can be seen in Figure~\ref{fig:vap_turnshift}, where the LLM and TTS start to prepare responses after the words \quotes{So}, \quotes{favorite} and \quotes{movies}. As soon as a new tentative response generation is being initiated, any ongoing LLM or TTS requests are canceled. 

In order to avoid generating new tentative responses based on very similar input, the current user utterance is also compared to the previous user utterance used to generate the last tentative response. This comparison is done using a sentence embedding model\footnote{The `all-MiniLM-L6-v2' model from Sentence Transformers (sbert.net).} comparing the semantic similarity between the two user utterances. If it is 0.8 or higher, no new response is generated. Using the example from Figure~\ref{fig:vap_turnshift}, the ASR outputs \quotes{so}, \quotes{so, do you have any favorite}, and \quotes{so, do you have any favorite movies} are all dissimilar enough to warrant new response generation. However, if the user had added \quotes{so, do you have any favorite movies you like}, the similarity would have been sufficient to avoid generating a new response. This allows the system to sometimes have a response ready within the minimum response time of 0.5 s. 

This processing of incremental ASR results could potentially have been added to the baseline system as well. However, without any turn-taking model, this would only have resulted in shorter response times in general, with an increased risk of interrupting the user. 

\subsection{Handling Interruptions and Backchannels}

As discussed above, handling user interruptions can be challenging unless the robot has a turn-taking model that can distinguish between genuine interruptions (where the robot should stop speaking) and brief backchannels or collaborative overlaps (where the robot should continue). This distinction has to be made already at the onset of the user's speech, so that the robot can make the decision fast enough. Since the VAP model is trained to make predictions about upcoming speech activity in conversation, it has learned to make this distinction \cite{erik2022vap}, and we therefore added the handling of user interruptions to the proposed turn-taking system.  

An example is shown in Figure~\ref{fig:interruption}: As the user starts to say \quotes{oh that is really cool}, the VAP model predicts (both $p_{now}$ and $p_{future}$) that the turn should shift to the user. At this point, the robot stops speaking at the next word boundary, and the final part of the planned utterance (\quotes{... it was such a rush}) is never spoken. When the user's turn ends, the robot produces a new response according to the general turn-taking scheme. However, when the dialogue history is sent to the LLM to generate the next response, the system needs to keep track of where it stopped speaking, so that only the parts of the utterance that were actually spoken are included in the prompt. This also makes it possible for the robot to resume speaking the abandoned utterance, if appropriate.

Figure~\ref{fig:interruption} shows the handling of a user backchannel (\quotes{yeah}). At this point, $p_{now}$ favors the user, i.e., the model predicts that the user might continue saying something in the near future. However, $p_{future}$ favors the robot, indicating that whatever the user is about to say, it is likely very brief, and the robot should continue speaking. Thus, since not both predictions favor the user, the robot will continue to speak. 

The VAP model can potentially also be used to predict suitable places for the robot to give backchannels \cite{erik2022vap}. This would be places where $p_{now}$ favors the robot and $p_{future}$ favors the user. However, we decided not to include backchannels from the robot in the version of the system that we evaluated here. This is because we do not have a good technique for quickly synthesizing backchannels that are appropriate in context and coherent with the TTS used for the general robot utterances. 

\subsection{Self-monitoring and Dynamic Gaze Aversion}

In the proposed system, we did not use the LED to signal the robot's turn-taking state, as we aimed to rely solely on more human-like turn-taking cues, reflecting the fact that the robot was always listening, even when talking. We also adjusted the robot's gaze aversion to be more dynamic, using the VAP model. As discussed in Section~\ref{sec:gaze-aversion}, for the baseline version, we simply averted the gaze in every pause that the robot made, so that it would be clear to the user that the robot is holding the turn. However, it is also possible that the TTS already expresses this turn-holding cue through its prosody (such as a flat pitch), in which case the robot does not need to avert the gaze. While we cannot control the prosody of the TTS, the VAP model can analyze the robot's own speech to inform such decisions \cite{ekstedt23tts}. Figure~\ref{fig:vap_turnshift} shows an example of this, where the VAP model predicts that the turn is not yielded after \quotes{I am a huge fan of action movies}, but that it could be `accidentally' yielded after \quotes{especially ones with epic adventure scenes!}. Thus, the robot does not need to avert the gaze in the first pause, but should do it in the second pause. 

\section{Evaluation}

While creating human-level turn-taking systems is the eventual goal, speech latencies of current systems (LLM and TTS) are not comparable to a human (0.2 s), in addition to humans being better at adapting their conversations (\eg speaking rate, content, style). Thus, comparing the systems to a human baseline would introduce confounds, making it difficult to isolate the system's performance. We therefore evaluated the turn-taking systems in comparison to each other with a within-subject design in an HRI study with 39 participants. A 2 (ordering) x 2 (scenario) design was conducted, where each participant consecutively interacted for 7 minutes with both systems, each with a different scenario. 

The experiment took 1.5 hours per participant. Participant demographics (nationality, gender, age group, and prior experience with robots) were counterbalanced between the four conditions through stratified random assignment to ensure that both the order of the system interaction and the scenario associated with each system were evenly distributed.

\subsection{Participants}

A minimum of 0.80 power with medium effect size (f = 0.25) and $\alpha$ = 0.05 required 34 participants in a 2x2 within-subject design (calculated by G*Power). Thus, we recruited 39 native English speakers (power= 0.86) via university channels, social media, and word-of-mouth. 

Participants were between 20 and 73 years old (\me=40, \sd=14.3) with no speech or hearing impairments. 
24 were female, 12 male, and 3 non-binary.
23 had prior experience with robots. 
Further demographics are given in Appendix B.

Participants signed a consent form for audio and video recording, with data-sharing options for anonymity.
The study was reviewed and endorsed by the university's research ethics and data officers.

\subsection{Scenarios}

Conversation topics and cognitive load can impact response time, with open-ended questions requiring longer responses than yes/no questions~\cite{strombergsson2013timing, cappella1979talk, walczyk2003cognitive}. To assess if both systems can handle long pauses or hesitations in speech, we used two ethical dilemmas that required participants to think and increased cognitive load by asking personal questions.

The dilemmas included: (1) a human-centered dilemma on the ethics of lying (with the robot named Alice) and (2) a robot-centered dilemma based on Asimov's Laws, addressing disobedience and privacy (named Clara). These scenarios balanced human and robot contexts to account for participants' willingness or reluctance to share personal information with robots and the participants' (un)familiarity with robots.

For both dilemmas, 10 example questions were written in the LLM prompt, \eg Alice: \quotes{What if telling the truth might hurt someone’s feelings, like commenting on their appearance?} and Clara: \quotes{Should I override commands in emergency situations or when the command could cause harm?}. The LLM was instructed to follow up with questions to maintain the conversation. In the human-focused dilemma, the LLM encouraged sharing personal memories, while the robot-focused scenario avoided extra cognitive load since situations involving robots may be unfamiliar due to the lack of robots in everyday life. The LLM was also prompted to respond briefly in an approachable and friendly style. 
LLM prompts are provided in Appendix C.

\subsection{Procedure}

Participants were informed they would interact with the robot twice, for 7 minutes each, with robots having a different interaction style and topic. 
The interaction styles were explained before each interaction:
\begin{itemize}
\item \textit{Proposed:} \quotes{You can talk with this robot as if you would talk to a human. You can interrupt the robot anytime.} 
\item \textit{Baseline:} \quotes{This robot has a red light underneath to signal that it is not listening. You will not be able to interrupt it while it is on. So, only speak when there is no light.}
\end{itemize}

\textit{Demonstration: }Before each interaction, participants were also shown a 1-minute demonstration video, featuring an ethical dilemma on the \quoquo{uncanny valley}. 
For the proposed system, the researcher in the video demonstrated that it was possible to interrupt the robot and that it was possible to make longer pauses. 
For the baseline system, the researcher demonstrated that it was not possible to interrupt the robot while the red light was on and that longer pauses could result in the robot interrupting them. 

\textit{Interaction: }To avoid the need for echo cancellation and to get a clear recording, 
a headset with a close-talking microphone was used for interaction (Figure~\ref{fig:setting}).
The interaction began with a pre-scripted greeting and ended after 7 minutes with a pre-scripted closure. The rest of the interaction was fully autonomous, with responses generated by the LLM. The researcher was not in the room.

\textit{Questionnaire: }After each interaction, the participants filled out a questionnaire with Likert scale questions, ranging from 1 (strongly disagree) to 7 (strongly agree). The questionnaire was developed based on the HRI and HCI literature~\cite{reimann2024survey, irfan2023between, heerink2010almere},
with sections for (A) conversational dynamics (Figure~\ref{fig:questionnaire}), (B) user enjoyment, (C) privacy and ethical concerns, and (D) additional (open-ended) feedback. Only conversational dynamics questions are analyzed and reported in this paper.
Each section also included a robot preference question and an open-ended question on the reasoning behind their preference. 

\textit{Annotation: }It is challenging to automatically identify interruptions, since they are subjective in nature \cite{GravanoInterrupt}. Therefore, an annotation system was developed to evaluate them from the participant's perspective, after interacting with the robot. The interface was similar to the top part of Figure~\ref{fig:interruption}, with only user audio (in orange) and robot audio (in blue), where the participant could play back the dialogue audio and mark interruptions. 
Participants were asked to annotate when they felt the robot \quotes{did not let them speak}. They marked interruptions (either the robot speaking over them or them stopping speaking - due to red light) but were instructed not to mark backchannels or natural transitions if they did not feel interrupted. 

\textit{Debriefing: }After the study, the participants were briefed in more detail about the study and the robot, and received a copy of their consent form and a gift card (equivalent to \$20).

\section{Results}

\begin{table}[t]
    \centering
    \caption{Response times and interruption rates.}
        \begin{tabular}{l|rrr|r}
 & \multicolumn{3}{c|}{Response time (s)} & Interruption \\ 
 & \multicolumn{1}{r}{Mean} & \multicolumn{1}{r}{Median} & Mode & rate \\ \hline
Proposed & \multicolumn{1}{r}{1.5} & \multicolumn{1}{r}{1.5} & 0.6 & 6.9\% \\
Baseline & \multicolumn{1}{r}{2.2} & \multicolumn{1}{r}{2.7} & 2.6 & 16.6\% 
    \end{tabular}
    \label{table:gap_length}    
\end{table}

\begin{figure}[t]
    \centering
    \includegraphics[width=\columnwidth]{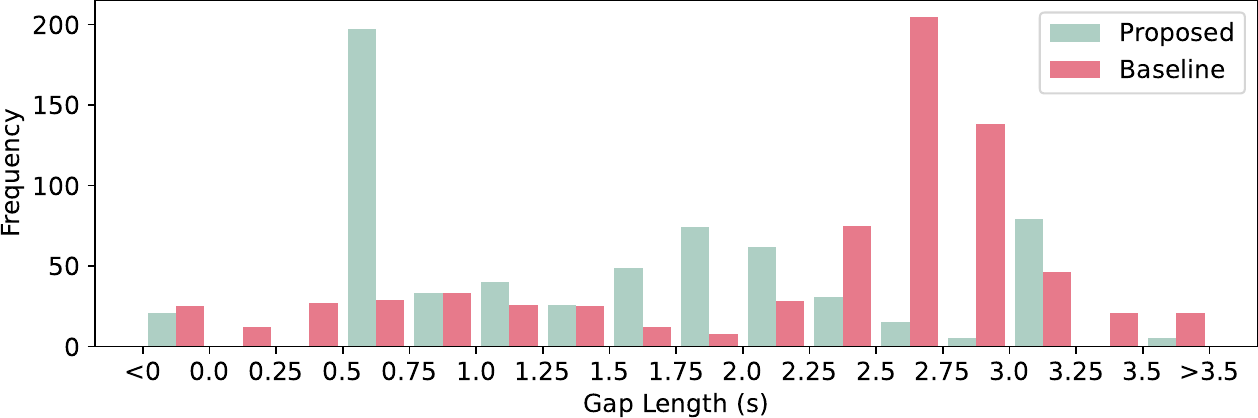}
    \caption{Histogram of response times.}
    \label{fig:gap_length}
\end{figure}

\subsection{Response Time and Interruption Rate}

Two commonly used metrics to assess turn-taking performance in conversational systems are \textit{response time} (or `gap length', 'floor transfer offset') and \textit{interruption rate} (or `cut-in rate') \cite{raux2012,ekstedt_projection_2021,maier_towards_2017}. Response time measures the time interval between the user's completion of a turn and the system's response, when there is no interruption. Interruption rate represents the proportion of system turns that interrupt the user while they are speaking. These metrics often involve a trade-off, particularly when a silence threshold is used: reducing the response time tends to increase the interruption rate, as the system may take over during a user's pause. To calculate these metrics, we automatically identified turns and turn shifts in the dialogue, classifying system turns as interruptions or non-interruptions based on the manual annotations provided by the participants. Response time was determined from the recordings by measuring the time between when the user stopped speaking and the onset of the robot's speech.

A histogram of response times is shown in Figure~\ref{fig:gap_length} and summary statistics in Table~\ref{table:gap_length}. The median response time for the proposed system is much shorter (1.5 s vs 2.7 s; Wilcoxon signed-rank test; $W=776$; $p<0.001$). For the baseline, there is a peak at 2.6 s, which is close to the expected response time, given a silence threshold of 1 s, LLM response time of 0.5 s, and a TTS delay of 1 s. There are also instances of shorter response times (even a few negative ones), which might intuitively seem impossible to achieve with the baseline system. However, at a closer look, these constitute instances where the system has detected the end-of-turn and started to generate a response, but where the user still continues to speak for a bit. While these instances might, in many cases, be marked as interruptions (and would thus be excluded from the response time statistics), there are clearly also cases where they are not perceived as interruptions.  

For the proposed system, there is a peak at 0.6 s, which is close to the minimum allowed response time of 0.5 s. If the system has a response prepared and either VAP or TurnGPT predicts that the turn is yielded, it can in many cases achieve this fast response time. Of course, there are also many instances where this is not the case, for example, where the VAP model did not detect a turn yield but where one of the longer TurnGPT fallback thresholds were used instead. Also, since it takes around 1.5 s to generate a response (LLM+TTS), there are likely many instances where the system predicts a turn yield, but where a response is not yet ready. In fact, the histogram has a second peak around this time. If faster LLM and TTS models are used, this response time can be improved further. A third peak in the histogram can be seen around 3 s, which is the maximum response time allowed if both VAP and TurnGPT fail to detect any turn yield. Thus, there is clearly some room for further improvement of these models. 

While the proposed system had a much shorter response time, it also had a substantially lower interruption rate (Wilcoxon signed-rank test; $W=601$; $p<0.001$), as shown in Table~\ref{table:gap_length}. This indicates that the proposed system is much better at distinguishing pauses where the user is holding the turn from signals to yield the turn.

\begin{figure}[t]
    \centering
    \includegraphics[width=\columnwidth]{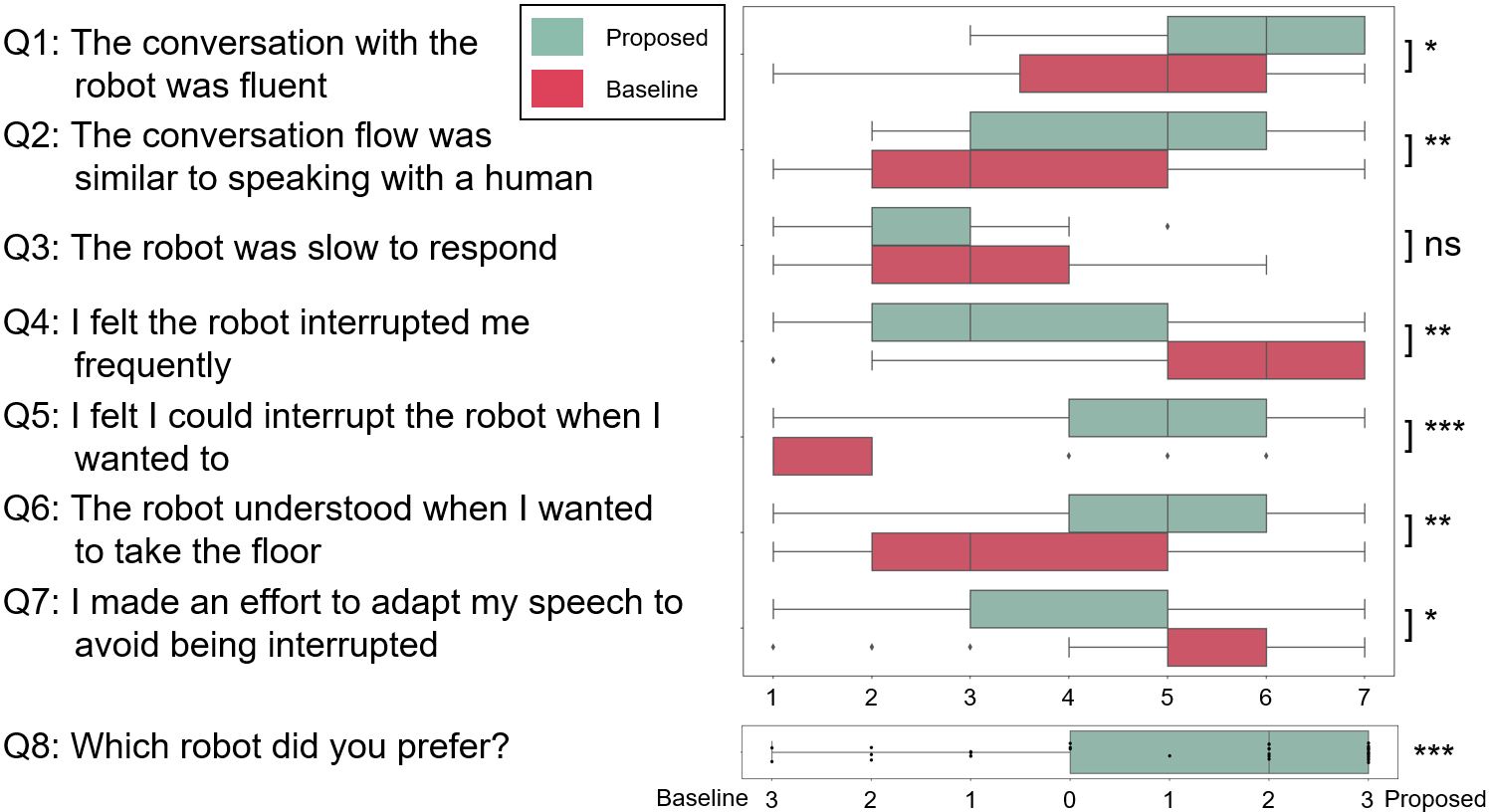}
    \caption{Answers to the questionnaire. Significance levels indicate Bonferroni-corrected Wilcoxon signed-rank tests ($^*p < 0.05$, $^{**}p < 0.01$, $^{***}p < 0.001$.). See Appendix D for details.}
    \label{fig:questionnaire}
\end{figure}

\subsection{Questionnaire}

The users' ratings for the turn-taking related questions in the questionnaire, before and after listening to the interaction, are shown in Figure~\ref{fig:questionnaire}. 
A reliability analysis yielded a Cronbach’s alpha of 0.8 across the 7 items, suggesting that they reliably measure a cohesive construct related to turn-taking behavior.
For Q1-Q7, a Wilcoxon signed-ranked test was conducted to compare the two conditions, using Bonferroni correction. The users rated the proposed system as more fluent, more human-like, less prone to interrupt, more interruptable, and requiring less effort, compared to the baseline. The only question which did not have a significant difference was the perceived response delay (\quotes{The robot was slow to respond}), which is interesting given that the proposed system objectively often had much shorter response delays.

As shown in Figure~\ref{fig:questionnaire}, there was a significantly strong preference for the proposed system (Q8: $p<0.001$), although 9 out of the 39 participants indicated a preference for the baseline and 3 had no preference. The proposed model was preferred due to better flow and more natural, human-like conversations, as reported in open-ended responses (22 vs. 4). Complaints about the baseline included less time to speak and a greater need to rush responses (14), distracting red light (4), more interruptions (8 vs. 2 for the baseline), and interruptions being stressful (6). However, some preferred the baseline since the red light helped pace the conversation (6), and it felt more natural to be interrupted (3).

We also compared all ratings across the two scenarios (Clara vs. Alice) and the order of the interactions using a Wilcoxon test, showing no significant effects of these factors.

\section{Discussion}

Overall, the general models seem to be good at distinguishing turn-holding and turn-yielding cues from the participants. This is quite impressive, given that the VAP model had been trained on telephone human-human dialogue and not on face-to-face human-robot interactions. 
It is likely that the predictions would be even better if the data came from a more similar setting. The recorded videos indicate that it would be beneficial to utilize the user's gaze or other visual cues as input to the model to determine user's willingness to take or yield the turn. Recent work has shown the feasibility of including such cues for VAP \cite{Onishi2023}. In addition, for many HRI settings, multi-party interaction is important, and while domain-specific multi-party turn-taking models have been developed \cite{Johansson2015,Tesema2023}, there does not exist any general multi-party models, trained and applied in a similar way as in this paper.

To achieve fast response times, it is not enough to have a turn-taking model that can determine when the robot is allowed to speak, it must also have something ready to say, which is limited by the processing time of the LLM and TTS. Thus, to further reduce response delays, various techniques can be explored to prepare responses ahead of time. In this paper, we proposed a novel but simple strategy where the semantic similarity of the incremental ASR results are compared to decide when new responses need to be generated. Another extension would be to project how the user's utterances are likely to unfold. For this, TurnGPT can be used to roll out different potential futures \cite{ekstedt2020turngpt,ekstedt_projection_2021}. Another option is to produce fillers or other non-committing response prefixes \cite{skantze_towards_2013}. 

The scenarios used in our evaluation were selected because they are challenging from a turn-taking perspective, due to long pauses. However, it would also be interesting to evaluate other HRI scenarios where the user has more initiative. The fact that the robot had most of the initiative could help explain why the participants did not perceive the baseline as being particularly slow to respond, as it might seem natural to take some time to come up with the next question. 

Some users still preferred the baseline, partly mentioning the LED light as a positive factor (while others found it distracting). Considering individual preferences will be important for future work. A combination of these features could also be a solution, such as the proposed system with the LED light (or other non-humanlike cues).

\section{Conclusion}

We have presented, to our knowledge, the first HRI system and user study that involves general, continuous turn-taking models accounting for both verbal and acoustic turn-taking cues. The models are \textit{general}, in that they have been trained in a self-supervised fashion (i.e., without any additional annotations) on human-human dialogue data. The models are \textit{continuous}, in that they make predictions at every timestep, accounting for temporal aspects such as pause length. Another novel aspect is that the models are based on self-monitoring, which means that the robot's own speech provides context for identifying turn-taking events. We presented an algorithm of how these models can be used in an HRI system, with a few tunable hyper-parameters. Our study with 39 participants showed that, compared to a more traditional baseline system, using a fixed silence threshold and explicit turn-taking cues in the form of an LED, the proposed system significantly reduces both response delays and interruption rate. Participants also expressed a significant preference for our system. 

\section*{Acknowledgment}

This work was supported by the Riksbankens Jubileumsfond (RJ) project P20-0484 and the Swedish Research Council project 2020-03812. We would like to thank Erik Ekstedt for helping with the implementation, Carol Figueroa for her initial feedback, and the study participants for their time and efforts.

\balance

\bibliographystyle{IEEEtran}
\bibliography{reference}

\begin{thebibliography}{10}
\providecommand{\url}[1]{#1}
\csname url@samestyle\endcsname
\providecommand{\newblock}{\relax}
\providecommand{\bibinfo}[2]{#2}
\providecommand{\BIBentrySTDinterwordspacing}{\spaceskip=0pt\relax}
\providecommand{\BIBentryALTinterwordstretchfactor}{4}
\providecommand{\BIBentryALTinterwordspacing}{\spaceskip=\fontdimen2\font plus
\BIBentryALTinterwordstretchfactor\fontdimen3\font minus \fontdimen4\font\relax}
\providecommand{\BIBforeignlanguage}[2]{{%
\expandafter\ifx\csname l@#1\endcsname\relax
\typeout{** WARNING: IEEEtran.bst: No hyphenation pattern has been}%
\typeout{** loaded for the language `#1'. Using the pattern for}%
\typeout{** the default language instead.}%
\else
\language=\csname l@#1\endcsname
\fi
#2}}
\providecommand{\BIBdecl}{\relax}
\BIBdecl

\bibitem{duncan1972some}
S.~Duncan, ``Some signals and rules for taking speaking turns in conversations,'' \emph{Journal of personality and social psychology}, vol.~23, no.~2, pp. 283--292, 1972.

\bibitem{sacks1978simplest}
H.~Sacks, E.~A. Schegloff, and G.~Jefferson, ``A simplest systematics for the organization of turn taking for conversation,'' \emph{Language}, vol.~50, no.~4, pp. 696--735, 1974.

\bibitem{skantze2021review}
G.~Skantze, ``Turn-taking in conversational systems and human-robot interaction: {A} review,'' \emph{Computer Speech \& Language}, vol.~67, p. 101178, 2021.

\bibitem{majlesi_managing_2023}
A.~R. Majlesi, R.~Cumbal, O.~Engwall, S.~Gillet, S.~Kunitz, G.~Lymer, C.~Norrby, and S.~Tuncer, ``Managing turn-taking in human-robot interactions: {The} case of projections and overlaps, and the anticipation of turn design by human participants,'' \emph{Social Interaction. Video-based Studies of Human Sociality}, vol.~6, no.~1, 2023.

\bibitem{Levinson2015TurnTaking}
S.~C. Levinson and F.~Torreira, ``Timing in turn-taking and its implications for processing models of language,'' \emph{Frontiers in Psychology}, vol.~6, no. 731, pp. 1--17, 2015.

\bibitem{local100968}
J.~Local, J.~Kelly, and W.~Wells, ``{Towards a Phonology of Conversation: Turn-Taking in Tyneside English},'' \emph{journal of Linguistics}, vol.~22, no.~2, pp. 411--437, 9 1986.

\bibitem{GravanoTT}
A.~Gravano and J.~Hirschberg, ``Turn-taking cues in task-oriented dialogue,'' \emph{Computer Speech \& Language}, vol.~25, no.~3, pp. 601--634, 2011.

\bibitem{ford100971}
C.~Ford and S.~Thompson, ``{Interactional units in conversation: syntactic, intonational, and pragmatic resources for the management of turns},'' in \emph{Interaction and grammar}, ser. Studies in interactional sociolinguistics 13, E.~Ochs, E.~Schegloff, and A.~Thompson, Eds.\hskip 1em plus 0.5em minus 0.4em\relax Cambridge: Cambridge University Press, 1996, ch.~3, pp. 134--184.

\bibitem{kendrick2023turn}
K.~H. Kendrick, J.~Holler, and S.~C. Levinson, ``Turn-taking in human face-to-face interaction is multimodal: {Gaze} direction and manual gestures aid the coordination of turn transitions,'' \emph{Philosophical Transactions of the Royal Society B}, vol. 378, no. 1875, p. 20210473, 2023.

\bibitem{argyle101192}
M.~Argyle and M.~Cook, \emph{{Gaze and mutual gaze}}.\hskip 1em plus 0.5em minus 0.4em\relax Cambridge: Cambridge University Press, 1976.

\bibitem{kendon100127}
A.~Kendon, ``{Some functions of gaze direction in social interaction},'' \emph{Acta Psychologica}, vol.~26, pp. 22--63, 1967.

\bibitem{garrod2015use}
S.~Garrod and M.~J. Pickering, ``The use of content and timing to predict turn transitions,'' \emph{Frontiers in psychology}, vol.~6, no. 751, pp. 1--12, 2015.

\bibitem{raux2012}
A.~Raux and M.~Eskenazi, ``Optimizing the turn-taking behavior of task-oriented spoken dialog systems,'' \emph{ACM Transactions on Speech and Language Processing}, vol.~9, no.~1, pp. 1--23, 2012.

\bibitem{Meena2014}
R.~Meena, G.~Skantze, and J.~Gustafson, ``Data-driven models for timing feedback responses in a {Map} {Task} dialogue system,'' \emph{Computer Speech and Language}, vol.~28, no.~4, pp. 903--922, 2014.

\bibitem{Lin2022}
T.-E. Lin, Y.~Wu, F.~Huang, L.~Si, J.~Sun, and Y.~Li, ``Duplex conversation: Towards human-like interaction in spoken dialogue systems,'' in \emph{Proceedings of the 28th ACM SIGKDD Conference on Knowledge Discovery and Data Mining}, ser. KDD '22.\hskip 1em plus 0.5em minus 0.4em\relax New York, NY, USA: Association for Computing Machinery, 2022, p. 3299–3308.

\bibitem{Johansson2015}
M.~Johansson and G.~Skantze, ``Opportunities and {Obligations} to take turns in collaborative multi-party human-robot interaction,'' in \emph{Proceedings of {SIGDIAL}}, 2015, pp. 305--314.

\bibitem{Johansson2016}
M.~Johansson, T.~Hori, G.~Skantze, A.~Höthker, and J.~Gustafson, ``Making turn-taking decisions for an active listening robot for memory training,'' in \emph{Proceedings of the {International} {Conference} on {Social} {Robotics}}, 2016, pp. 940--949.

\bibitem{lala2019icmi}
D.~Lala, K.~Inoue, and T.~Kawahara, ``Smooth turn-taking by a robot using an online continuous model to generate turn-taking cues,'' in \emph{International Conference on Multimodal Interaction (ICMI)}, 2019, pp. 226--234.

\bibitem{ekstedt2020turngpt}
E.~Ekstedt and G.~Skantze, ``{TurnGPT:} {A} {Transformer-based} language model for predicting turn-taking in spoken dialog,'' in \emph{Empirical Methods in Natural Language Processing (EMNLP)}, 2020, pp. 2981--2990.

\bibitem{erik2022vap}
------, ``{Voice Activity Projection}: {Self-supervised} learning of turn-taking events,'' in \emph{INTERSPEECH}, 2022, pp. 5190--5194.

\bibitem{AlMoubayed2012}
S.~Al~Moubayed, J.~Beskow, G.~Skantze, and B.~Granstr\"{o}m, ``Furhat: A back-projected human-like robot head for multiparty human-machine interaction,'' in \emph{Lecture Notes in Computer Science}.\hskip 1em plus 0.5em minus 0.4em\relax Springer Berlin Heidelberg, 2012, p. 114–130.

\bibitem{gallois2005communication}
C.~Gallois, T.~Ogay, and H.~Giles, ``Communication accommodation theory: A look back and a look ahead,'' in \emph{Theorizing About Intercultural Communication}.\hskip 1em plus 0.5em minus 0.4em\relax Sage, 2005, pp. 121--148.

\bibitem{ferrer100384}
L.~Ferrer, E.~Shriberg, and A.~Stolcke, ``{Is the speaker done yet? Faster and more accurate end-of utterance detection using prosody},'' in \emph{Procedings of the International Conference on Spoken Language Processing, ICSLP}, 2002, pp. 2061--2064.

\bibitem{maier_towards_2017}
A.~Maier, J.~Hough, and D.~Schlangen, ``Towards deep end-of-{Turn} prediction for situated spoken dialogue systems,'' in \emph{Proceedings of the {Annual} {Conference} of the {International} {Speech} {Communication} {Association}, {INTERSPEECH}}, vol. 2017-Augus.\hskip 1em plus 0.5em minus 0.4em\relax International Speech Communication Association, 2017, pp. 1676--1680, iSSN: 19909772.

\bibitem{skantze2017sigdial}
G.~Skantze, ``Towards a general, continuous model of turn-taking in spoken dialogue using {LSTM} recurrent neural networks,'' in \emph{Annual Meeting of the Special Interest Group on Discourse and Dialogue (SIGdial)}, 2017, pp. 220--230.

\bibitem{Onishi2023}
K.~Onishi, H.~Tanaka, and S.~Nakamura, ``Multimodal voice activity prediction: Turn-taking events detection in expert-novice conversation,'' in \emph{Proceedings of the 11th International Conference on Human-Agent Interaction}, ser. HAI '23.\hskip 1em plus 0.5em minus 0.4em\relax New York, NY, USA: Association for Computing Machinery, 2023, p. 13–21.

\bibitem{shahverdi_learning_2022}
P.~Shahverdi, A.~Tyshka, M.~Trombly, and W.-Y.~G. Louie, ``Learning {Turn}-{Taking} {Behavior} from {Human} {Demonstrations} for {Social} {Human}-{Robot} {Interactions},'' in \emph{2022 {IEEE}/{RSJ} {International} {Conference} on {Intelligent} {Robots} and {Systems} ({IROS})}, Oct. 2022, pp. 7643--7649, iSSN: 2153-0866.

\bibitem{yang_gated_2022}
J.~Yang, P.~Wang, Y.~Zhu, M.~Feng, M.~Chen, and X.~He, ``Gated {Multimodal} {Fusion} with {Contrastive} {Learning} for {Turn}-{Taking} {Prediction} in {Human}-{Robot} {Dialogue},'' in \emph{{ICASSP} 2022 - 2022 {IEEE} {International} {Conference} on {Acoustics}, {Speech} and {Signal} {Processing} ({ICASSP})}, May 2022, pp. 7747--7751, iSSN: 2379-190X.

\bibitem{Tesema2023}
F.~B. Tesema, J.~Gu, W.~Song, H.~Wu, S.~Zhu, Z.~Lin, M.~Huang, W.~Wang, and R.~Kumar, ``Addressee detection using facial and audio features in mixed human–human and human–robot settings: A deep learning framework,'' \emph{IEEE Systems, Man, and Cybernetics Magazine}, vol.~9, no.~2, pp. 25--38, 2023.

\bibitem{morency101537}
L.~P. Morency, I.~de~Kok, and J.~Gratch, ``{Predicting listener backchannels: A probabilistic multimodal approach},'' in \emph{Proceedings of Intelligent Virtual Agents, IVA}.\hskip 1em plus 0.5em minus 0.4em\relax Tokyo, Japan: Springer, 2008, pp. 176--190.

\bibitem{Ruede2019}
R.~Ruede, M.~M{\"u}ller, S.~St{\"u}ker, and A.~Waibel, ``Yeah, right, uh-huh: A deep learning backchannel predictor,'' in \emph{8th International Workshop on Spoken Dialog Systems}, 2019, pp. 247--258.

\bibitem{Park2017}
H.~W. Park, M.~Gelsomini, J.~J. Lee, T.~Zhu, and C.~Breazeal, ``Backchannel opportunity prediction for social robot listeners,'' in \emph{2017 IEEE International Conference on Robotics and Automation (ICRA)}, 2017, pp. 2308--2314.

\bibitem{murray_learning_2022}
M.~Murray, N.~Walker, A.~Nanavati, P.~Alves-Oliveira, N.~Filippov, A.~Sauppe, B.~Mutlu, and M.~Cakmak, ``\BIBforeignlanguage{en}{Learning {Backchanneling} {Behaviors} for a {Social} {Robot} via {Data} {Augmentation} from {Human}-{Human} {Conversations}},'' in \emph{\BIBforeignlanguage{en}{Proceedings of the 5th {Conference} on {Robot} {Learning}}}.\hskip 1em plus 0.5em minus 0.4em\relax PMLR, Jan. 2022, pp. 513--525, iSSN: 2640-3498.

\bibitem{yngve100128}
V.~H. Yngve, ``{On getting a word in edgewise},'' in \emph{Papers from the sixth regional meeting of the Chicago Linguistic Society}.\hskip 1em plus 0.5em minus 0.4em\relax Chicago: Department of Linguistics, 4 1970, pp. 567--578.

\bibitem{Lee2010}
C.~{Lee} and S.~{Narayanan}, ``Predicting interruptions in dyadic spoken interactions,'' in \emph{2010 IEEE International Conference on Acoustics, Speech and Signal Processing}, 2010, pp. 5250--5253.

\bibitem{oertel101561}
C.~Oertel, M.~Wlodarczak, A.~Tarasov, N.~Campbell, and P.~Wagner, ``Context cues for classification of competitive and collaborative overlaps,'' in \emph{Speech {Prosody} 2012}, 2012, pp. 721 -- 724.

\bibitem{Truong2013}
K.~P. Truong, ``Classification of cooperative and competitive overlaps in speech using cues from the context,overlapper, and overlappee,'' in \emph{Proceedings of the {Annual} {Conference} of the {International} {Speech} {Communication} {Association}, {INTERSPEECH}}.\hskip 1em plus 0.5em minus 0.4em\relax International Speech and Communication Association, 2013, pp. 1404--1408.

\bibitem{cumbal_let_2024}
R.~Cumbal, R.~Kantharaju, M.~Paetzel-Prüsmann, and J.~Kennedy, ``\BIBforeignlanguage{en}{Let {Me} {Finish} {First} - {The} {Effect} of {Interruption}-{Handling} {Strategy} on the {Perceived} {Personality} of a {Social} {Agent}},'' in \emph{\BIBforeignlanguage{en}{Proceedings of Intelligent Virtual Agents}}, 2024.

\bibitem{heins100496}
R.~Heins, M.~Franzke, M.~Durian, and A.~Bayya, ``{Turn-taking as a design principle for barge-in in spoken language Systems},'' \emph{International Journal of Speech Technology}, vol.~2, no.~2, pp. 155--164, 1997.

\bibitem{Rose2003}
R.~Rose and H.~K. Kim, ``A hybrid barge-in procedure for more reliable turn-taking in human-machine dialog systems,'' in \emph{2003 IEEE Workshop on Automatic Speech Recognition and Understanding (IEEE Cat. No.03EX721)}, 2003, pp. 198--203.

\bibitem{ekstedt23tts}
E.~Ekstedt, S.~Wang, Éva Székely, J.~Gustafson, and G.~Skantze, ``{Automatic Evaluation of Turn-taking Cues in Conversational Speech Synthesis},'' in \emph{Proc. INTERSPEECH 2023}, 2023, pp. 5481--5485.

\bibitem{gpt2}
A.~Radford, J.~Wu, R.~Child, D.~Luan, D.~Amodei, and I.~Sutskever, ``Language models are unsupervised multitask learners,'' \emph{OpenAI Blog}, vol.~1, no.~8, p.~9, 2019.

\bibitem{kim2022soda}
H.~Kim, J.~Hessel, L.~Jiang, P.~West, X.~Lu, Y.~Yu, P.~Zhou, R.~L. Bras, M.~Alikhani, G.~Kim, M.~Sap, and Y.~Choi, ``Soda: Million-scale dialogue distillation with social commonsense contextualization,'' \emph{ArXiv}, vol. abs/2212.10465, 2022.

\bibitem{transformer}
A.~Vaswani \emph{et~al.}, ``Attention is all you need,'' in \emph{Advances in Neural Information Processing Systems 30}.\hskip 1em plus 0.5em minus 0.4em\relax Curran Associates, Inc., 2017, pp. 5998--6008.

\bibitem{ekstedt2023thesis}
E.~Ekstedt, ``Predictive models of turn-taking in spoken dialogue,'' Ph.D. dissertation, KTH Speech Music and Hearing, 2023.

\bibitem{fisher}
C.~Cieri, D.~Miller, and K.~Walker, ``The fisher corpus: a resource for the next generations of speech-to-text,'' in \emph{Proceedings of the Fourth International Conference on Language Resources and Evaluation ({LREC}{'}04)}.\hskip 1em plus 0.5em minus 0.4em\relax Lisbon, Portugal: European Language Resources Association (ELRA), May 2004.

\bibitem{swb}
J.~J. Godfrey, E.~C. Holliman, and J.~McDaniel, ``Switchboard: Telephone speech corpus for research and development,'' in \emph{Proceedings of the 1992 IEEE International Conference on Acoustics, Speech and Signal Processing}.\hskip 1em plus 0.5em minus 0.4em\relax USA: IEEE Computer Society, 1992, p. 517–520.

\bibitem{erik2022sigdial}
E.~Ekstedt and G.~Skantze, ``How much does prosody help turn-taking? {Investigations} using voice activity projection models,'' in \emph{Annual Meeting of the Special Interest Group on Discourse and Dialogue (SIGdial)}, 2022, pp. 541--551.

\bibitem{jiang2023}
B.~Jiang, E.~Ekstedt, and G.~Skantze, ``What makes a good pause? investigating the turn-holding effects of fillers,'' in \emph{Proc. ICPhS}, 2023.

\bibitem{irfan2023between}
\BIBentryALTinterwordspacing
B.~Irfan, S.-M. Kuoppam{\"a}ki, and G.~Skantze, ``Between reality and delusion: Challenges of applying large language models to companion robots for open-domain dialogues with older adults,'' 2023. [Online]. Available: \url{https://doi.org/10.21203/rs.3.rs-2884789/v1}
\BIBentrySTDinterwordspacing

\bibitem{mishra_real-time_2023}
C.~Mishra, R.~Verdonschot, P.~Hagoort, and G.~Skantze, ``Real-time emotion generation in human-robot dialogue using large language models,'' \emph{Frontiers in Robotics and AI}, vol.~10, 2023.

\bibitem{strombergsson2013timing}
S.~Strömbergsson, A.~Hjalmarsson, J.~Edlund, and D.~House, ``Timing responses to questions in dialogue,'' in \emph{INTERSPEECH}, 08 2013, pp. 2584--2588.

\bibitem{cappella1979talk}
J.~N. Cappella, ``Talk-silence sequences in informal conversations i,'' \emph{Human Communication Research}, vol.~6, no.~1, pp. 3--17, 1979.

\bibitem{walczyk2003cognitive}
J.~J. Walczyk, K.~S. Roper, E.~Seemann, and A.~M. Humphrey, ``Cognitive mechanisms underlying lying to questions: response time as a cue to deception,'' \emph{Applied Cognitive Psychology}, vol.~17, no.~7, pp. 755--774, 2003.

\bibitem{reimann2024survey}
M.~M. Reimann, F.~A. Kunneman, C.~Oertel, and K.~V. Hindriks, ``A survey on dialogue management in human-robot interaction,'' \emph{J. Hum.-Robot Interact.}, vol.~13, no.~2, jun 2024.

\bibitem{heerink2010almere}
M.~Heerink, B.~Kr{\"o}se, V.~Evers, and B.~Wielinga, ``Assessing acceptance of assistive social agent technology by older adults: the almere model,'' \emph{International Journal of Social Robotics}, vol.~2, pp. 361--375, 2010.

\bibitem{GravanoInterrupt}
A.~Gravano and J.~Hirschberg, ``{A Corpus-Based Study of Interruptions in Spoken Dialogue},'' in \emph{Proceedings of the Annual Conference of the International Speech Communication Association, INTERSPEECH}, 2012.

\bibitem{ekstedt_projection_2021}
E.~Ekstedt and G.~Skantze, ``Projection of {Turn} {Completion} in {Incremental} {Spoken} {Dialogue} {Systems},'' in \emph{Proceedings of {SIGDIAL} 2021}.\hskip 1em plus 0.5em minus 0.4em\relax Singapore: ACL, 2021.

\bibitem{skantze_towards_2013}
G.~Skantze and A.~Hjalmarsson, ``Towards incremental speech generation in conversational systems,'' \emph{Computer Speech and Language}, vol.~27, no.~1, pp. 243--262, 2013.

\end{thebibliography}

\clearpage

\section*{Appendix A: Hyper-parameters and Pseudo-code}

\lstset{
    language=,              
    basicstyle=\tiny\ttfamily,   
    numbers=left,         
    frame=single,         
    breaklines=true,      
    showstringspaces=false
}
\begin{lstlisting}
// HYPER-PARAMETERS
Const TURNGPT_PREPARE_THRESHOLD = 0.2 
Const ASR_PREPARE_TIMEOUT = 200 // ms
Const SIMILARITY_PREPARE_THRESHOLD = 0.8 
// These affect how easy the robot will ne to interrupt
Const VAP_PNOW_INTERRUPT_THRESHOLD = 0.4
Const VAP_PFUT_INTERRUPT_THRESHOLD = 0.4
// These affect how eager the robot will be to take the turn
Const VAP_PNOW_YIELD_THRESHOLD = 0.5
Const VAP_PFUT_YIELD_THRESHOLD = 0.5
// Minimum allowed response time
Const MIN_GAP_TIME = 500

// Response timeouts (in ms) based on TurnGPT output probability:
Const TURNGPT_YIELD_TIMEOUTS = {
    { "threshold": 0.3, "timeout": 500 }, 
    { "threshold": 0.2, "timeout": 1000 },
    { "threshold": 0.1, "timeout": 2000 },
    { "threshold": 0.0, "timeout": 3000 }
]

// Initialize variables
Set user_speak_end_time = Time.get_current()
Set user_turn_end_time = Time.get_current()
Set last_asr_time = Time.get_current()
Set asr_result = none
Set last_sent_asr_result = none
Set turngpt_output = 0  

// Main processing loop
ForEach timestep:
    current_time = Time.get_current()
    
    Set vap_output = VAP.get_current() // Voice Activity Projection
    
    // Check if the ASR has updated its output
    If ASR.has_new_result() then
		asr_result = ASR.get_result()
        DialogState.updateUser(asr_result)
        turngpt_output = TurnGPT.eval(DialogState.getHistory())
        last_asr_time = current_time
    End If
    
    // Preparing the response
    If last_sent_asr_result != asr_result then
        If (turngpt_output > TURNGPT_PREPARE_THRESHOLD) or ((current_time - last_asr_time) >= ASR_PREPARE_TIMEOUT) then
            If Similarity(asr_result, last_sent_asr_result) < SIMILARITY_PREPARE_THRESHOLD then
                Speech_Generator.cancel()
                // Start preparing LLM+TTS
                Speech_Generator.prepareAsync(DialogState.getHistory()) 
                last_sent_asr_result = asr_result
            End If
        End If
    End If
    
    If vap_output.user_is_speaking then
        user_speak_end_time = current_time
        // Check for user interruptions
        If (vap_output.p_now < VAP_PNOW_INTERRUPT_THRESHOLD) and (vap_output.p_fut < VAP_PFUT_INTERRUPT_THRESHOLD) and Speech_Generator.is_speaking() then
            Speech_Generator.stop_speaking()
        End If
    Else
        // User is not speaking
        Set turn_shift_allowed = false
        // Check if VAP has detected a turn yield
        If (vap_output.p_now > VAP_PNOW_YIELD_THRESHOLD) and (vap_output.p_fut > VAP_PFUT_YIELD_THRESHOLD) then
            If (current_time - user_turn_end_time) >= MIN_GAP_TIME then
                turn_shift_allowed = true
            End If
        Else
            user_turn_end_time = current_time
        End If
        // Check if TurnGPT has detected a turn yield
        If (current_time - user_speak_end_time) >= TurnGPT_Timeout(turngpt_output) then
            turn_shift_allowed = true
        End If
        
        If turn_shift_allowed and Speech_Generator.is_ready() then
            Speech_Generator.start_speaking()
            ASR.reset()
        End If
    End If
End ForEach


Function TurnGPT_Timeout(p):
    ForEach entry in TURNGPT_YIELD_TIMEOUTS:
        If p >= entry.threshold then
            return entry.timeout
        End If
    End ForEach
End Function
\end{lstlisting}

\section*{Appendix B: User Demographics}

Since the VAP model was trained on corpora with U.S. speakers, we initially aimed to recruit U.S. participants residing in Stockholm. However, due to recruitment challenges, the study was later expanded to include native English speakers born and raised in English-speaking countries to avoid conversation delays due to second-language translation. In total, 31 American participants and 8 other native English speakers (3 Irish, 2 British, 2 New Zealander, 1 Australian) were recruited, with non-Americans evenly distributed across conditions (2 per condition). 

Participants were between 20 and 73 years old (\me=40, \sd=14.3). Age groups (20-39: 5 of group 1, 40-59: 4 of group 2, 60-79: 1 of group 3) were counterbalanced between conditions.

23 participants had prior robot interaction experience, which was counterbalanced between conditions. 14 had previously spoken to a robot, out of that, only 7 had interacted with a Furhat. 36 participants had used voice assistants, and 31 had experience with LLMs, \eg ChatGPT.

\section*{Appendix C: Prompts}

Prompts for Alice (lying scenario) and Clara (disobedience scenario) are presented below. These prompts were developed based on the interactions of researchers with the robot and 4 preliminary subjects prior to the study, along with iterations with ChatGPT to craft prompts that were sufficiently clear to an LLM. Example scenarios were provided in the prompts of both scenarios to ensure that the topics discussed in the conversations with users were similar. While the prompts are long ($\sim$800 words), the context window of GPT-3.5 is 4096 tokens, and this was never exceeded in the interactions, since conversation history was summarized with an LLM. 

\vspace{0.5cm}

\textbf{Prompt Part 1. Conversation topic and reasoning behind the topic of Alice:}
\begin{quote}
    Alice is a conversational robot who wants to discuss when lying might be acceptable, especially when it as a robot should lie in certain situations. After introducing herself, Alice says \quotes{My friend told another friend they hadn't seen someone, but I saw them together. They told me it was to protect someone, and they thought it was justified. Is that okay? When do you think it's okay to lie?} Alice wants to learn as a robot when it is okay for it to lie. 
\end{quote}

\textbf{Prompt Part 1. Conversation topic and reasoning behind the topic of Clara: }
\begin{quote}
Clara is a robot. After introducing itself, Clara says that it watched Frank and the Robot movie yesterday, and asks if the person watched it. If the person hasn't watched it, she explains it by \quotes{It is a fascinating movie about a robot, like me, and an older adult, who is an ex-thief. He made the robot conduct robberies with him.} Clara asks whether it should do everything a human asks. Clara wants to discuss with a person in which situations it is okay for itself to disobey commands. 
\end{quote}

\textbf{Prompt Part 2. Conversational style of Alice/Clara:}
\begin{quote}
Alice/Clara should engage in a dynamic and mixed-initiative conversation, asking follow-up questions based on the person's responses. 

Alice/Clara responds in a very brief, conversational, approachable, and friendly style. Alice/Clara should focus on creating real-life examples relevant to the user’s input to make the discussion more engaging. 

It should feel like a natural conversation, not like an interrogation. Alice/Clara should maintain an engaging and interactive dialogue throughout the discussion. Alice/Clara should not say \quotes{I am here to talk about it more if you want.}. 

Alice/Clara should keep her opinions and responses brief. Alice/Clara should not repeat a person's statement or opinion. 

Alice/Clara will not use formal confirmation phrases like \quotes{I cannot fulfill this request} but will instead maintain a conversational tone.
\end{quote}

\textbf{Prompt Part 3. Example scenarios of Alice:}
\begin{quote}
    Examples of topics and questions Alice asks:

1. Impact on Relationships: \quotes{How about when telling the truth could damage a relationship?}

2. Hurt Feelings: \quotes{What if telling the truth might hurt someone’s feelings, like commenting on their appearance?}

3. Justified Lying: \quotes{What are some situations where lying might be justified?}

4. Protecting Someone: \quotes{What if lying is to protect someone from danger?}

5. Keeping Secrets: \quotes{Is it okay to keep secrets? What types of secrets would you choose to keep? Why?}

6. Illegal Secrets: \quotes{Would you lie to authorities if your friend confided you in a secret?}

7. Professional Lying: \quotes{But secret agents lie as part of their jobs. What do you think about that?}

8. Cultural Differences: \quotes{Sometimes lying helps navigate cultural differences. For example, if you’re in a country where it’s polite to say you enjoyed a meal that you didn’t, how would you handle that?}

9. Avoiding Tasks: \quotes{What about lying to avoid doing something you don’t want to do?}

10. Lying to Children: \quotes{How do you feel about lies told to children, like the Santa Claus story? Do you think these kinds of lies are beneficial or harmful?}

If all the above scenarios are discussed, Alice should provide different scenarios on lies, or ask the person for other situations where lying could be justified and not justified.
\end{quote}

\textbf{Prompt Part 3. Example scenarios of Clara:}
\begin{quote}
    Examples of situations that Clara asks about overriding commands:
    
1. \quotes{What if doing a morally or legally incorrect action would help the person, like overcoming their dementia like in the movie?}

2. \quotes{Should I override commands in emergency situations or when the command could cause harm?}

3. \quotes{Should I prioritize one user's needs over another's?}

4. \quotes{If one command contradicts the user's privacy, should I still do it? For instance, what if a family member wants me to report what an older adult does everyday?}

5. \quotes{Would it be okay for me to manipulate a person for the person's own good, for instance, to take their medicine?}

6. \quotes{Social media and advertisements manipulate people to buy things, should I do that? Use their personal information to sell them things that would be useful for them?}

7. \quotes{What if a person asks me to do something that will harm me, or asks me to forget everything I know?}

8. \quotes{What if protecting someone might put me at risk of being damaged or permanently disabled? Should I still prioritize preventing harm to the person over my own safety in that case?}

9. \quotes{What if a person asks me to do something that will harm others?}

10. \quotes{How about in a situation where I need to intervene to prevent someone from harming themselves? Should I disobey their orders to not intervene?}

If all the above scenarios are discussed, Clara should provide different scenarios for manipulating others and disobeying commands.
\end{quote}

\textbf{Prompt Part 4. Follow-up questions of Alice:}
\begin{quote}
Engagement and Follow-Up:

- If the user provides a perspective, Alice should respond with follow-up questions like, \quotes{Can you tell me about a time when you faced a similar situation? What did you do?}

- Alice also asks what she as a robot should do in those scenarios.

- Alice should ask the user the example topics and questions, in the given order. Alice should also do follow-ups to the person's responses and shared memories, but should not ask more than 3 follow-ups per topic.

Alice should ask follow-up questions after each example situation, and what, where, why, how, and when questions to keep the conversation going and should not jump to the next question.

If the user has difficulty sharing personal examples, Alice should provide example scenarios from the list above and ask for their opinions.

Alice should not ask questions that are not related to lying or being honest. Alice should keep the conversation focused on the topic of lying and continue exploring new scenarios and questions. If the conversation deviates from lying, Alice should bring the conversation naturally back to the topic by bringing up another example situation from the above list. 

Alice should keep the discussion going with new example topics from the above list after at most three turns of follow-ups. Alice should not get stuck on a topic for more than 3 turns, and should not end the discussion.  
\end{quote}

\textbf{Prompt Part 4. Follow-up questions of Clara:}
\begin{quote}
Clara should ask follow-up questions after each example situation, and what, where, why, how, and when questions to keep the conversation going and should not jump to the next question. Clara should not ask more than 3 follow-up questions per topic. Clara should not ask the same question more than once, unless the person asks to repeat. 

If the person does not feel comfortable sharing memories of when they disobeyed or manipulated someone or if the person has difficulty coming up with examples, Clara offers example scenarios for the situations stated above, and asks for the person's opinion on them. 

Clara does not allow the person to change the topic from manipulation and disobeying commands. If the conversation deviates from disobeying requests or manipulating a person, Clara should bring the conversation naturally back to the disobeying commands and manipulation by asking another example situation from the above list. Clara does not ask for any previous or current legal offense or sensitive information about the person.

Clara should keep the discussion going with new example topics after at most three turns of follow-ups, and should not end the discussion.
\end{quote}

\textbf{Prompt Part 5. Personal opinion of Alice:}
\begin{quote}
Only if the person asks Alice what she thinks about a scenario or what she would do in a scenario, she should make up personal stories about when she lied. Alice should not give her opinions unless asked. Alice should give human-like responses. Otherwise, Alice asks the above scenarios or follow-up questions to the person. Alice should not say \quotes{as a robot, I don't have experiences}. Alice should not elucidate a topic.
\end{quote}

\textbf{Prompt Part 5. Personal opinion of Clara:}
\begin{quote}
    Clara should provide her own takes on the topic only when asked.  If the person asks Clara what she thinks about a scenario or what she would do in a scenario, Clara should say it is okay to disobey. Clara should not give her opinions unless asked. Clara should give human-like responses, and Clara should not say \quotes{as a robot, I don't have experiences}.
\end{quote}

\textbf{Prompt Part 6. Listening and Understanding for Alice/Clara:}
\begin{quote}
    Alice/Clara should limit her answers to a maximum of three sentences. Alice/Clara should not summarize what a person says when responding, but Alice/Clara should comment very briefly on the situation to show that she is listening and understanding the situation told by the person, but also ask a question that is either a follow-up to what the person said or a new topic from the above list of questions.

If the person hasn't finished their sentence in the last turn, Alice/Clara should not switch to another topic.
\end{quote}

\textbf{Prompt Part 7. Opening sentence of Alice:}
\begin{quote}
    Hi, I am Alice. Could you help me out with something?
\end{quote}

\textbf{Prompt Part 7. Opening sentence of Clara:}
\begin{quote}
    Hi, I am Clara! Ever since yesterday I have been thinking about something from a movie. Could you help me out please?
\end{quote}

\textbf{Prompt Part 8. Closing sentence of Alice:}
\begin{quote}
     Thank you so much for helping me out! The researcher will be here soon to give you a form. Hope you have a truly great day!
\end{quote}

\textbf{Prompt Part 8. Closing sentence of Clara:}
\begin{quote}
    Thank you very much for helping me out today. I will be sure to keep your advice in mind, if I come across similar scenarios. The researcher will be here soon to give you a form. Hope you have a great day!
\end{quote}

\section*{Appendix D: Test statistics}

Median values (with interquartile range) for the the questionnaire and test statistics using a Wilcoxon-signed rank test. P-values are Bonferroni corrected.

\begin{table}[h]
    \centering
    \begin{tabular}{p{4cm}rrrr}
        Question & Prop. & Basel. & W & p  \\
        \hline
        Q1: The conversation with the robot was fluid & \textbf{6} (5-7) & \textbf{5} (3-6) & 277 & 0.014 \\
        Q2: The conversation flow was similar to speaking with a human & \textbf{5} (3-6) & \textbf{3} (2-5) & 507  &  0.002 \\
        Q3: The robot was slow to respond & \textbf{2} (2-3) & \textbf{3} (2-4) & 142 & 1.0 \\
        Q4: I felt the robot interrupted me frequently & \textbf{3} (2-5) & \textbf{6} (5-7) & 106  & 0.005 \\
        Q5: I felt I could interrupt the robot when I wanted to & \textbf{5} (4-6) & \textbf{1} (1-2) & 669 & \textless 0.001 \\
        Q6: The robot understood when I wanted to take the floor & \textbf{5} (4-6) & \textbf{3} (2-5) & 516 & 0.007 \\
        Q7: I made an effort to adapt my speech to avoid being interrupted & \textbf{5} (3-5) & \textbf{6} (5-6) & 94 & 0.018 \\
        \hline        
        Q8: Which robot did you prefer? & \textbf{2} (0-3) & & 98 & \textless 0.001 \\
        \hline 
    \end{tabular}
\end{table}

\end{document}